\pdfoutput=1

\documentclass[11pt]{article}

\usepackage{EMNLP2022}

\usepackage{times}
\usepackage{latexsym}

\usepackage[T1]{fontenc}

\usepackage[utf8]{inputenc}

\usepackage{microtype}

\usepackage{inconsolata}

\usepackage{algorithm}
\usepackage{algpseudocode}
\usepackage{amsfonts}
\usepackage{amssymb,amsmath}
\usepackage[utf8]{inputenc}
\usepackage{xparse}
\NewDocumentCommand{\avi}
{ mO{} }{\textcolor{red}{\textsuperscript{\textit{Avi}}\textsf{\textbf{\small[#1]}}}}
\title{Zero-Shot Dynamic Quantization for Transformer Inference}

 \makeatletter
\newcommand{\printfnsymbol}[1]{%
  \textsuperscript{\@fnsymbol{#1}}%
}
\makeatother

\author{Yousef El-Kurdi\thanks{~~Equal contribution.}~~~~~~~Jerry Quinn\printfnsymbol{1}~~~~~~~Avirup Sil \\
   IBM Research AI\\ 
  \texttt{\{yousefelk, jlquinn, avi\}@us.ibm.com}}
  
\begin{document}
\maketitle
\begin{abstract}
We introduce a novel run-time method for significantly reducing the accuracy loss associated with quantizing BERT-like models to 8-bit integers.  
Existing methods for quantizing models either modify the training procedure, or they require an additional calibration step to adjust parameters that also requires a selected held-out dataset.
Our method permits taking advantage of quantization without the need for these adjustments.
We present results on several NLP tasks demonstrating the usefulness of this technique.

\end{abstract}

\section{Introduction}

Transformer-based Neural Networks (NN) such as BERT \cite{devlin2018bert}, RoBERTa \cite{liu2019roberta} and XLM-R \cite{conneau2019unsupervised}, pre-trained on large amounts of data, have led to state-of-the-art (SOTA) results on many NLP tasks such as machine translation \cite{zhu2019incorporating}, text classification \cite{DBLP:journals/corr/abs-1804-07461}
and question answering \cite{47761,DBLP:journals/corr/abs-2003-05002}. However, run-time inference of such large models is very costly due to their large computational requirements.
In addition, deploying these models on smaller footprint mobile devices \cite{ravi2021soda} or cost-effective \cite{sanh2019distilbert,jiao2020tinybert} CPU based machines require aggressive optimization techniques for both speed and network size.
One popular speed optimization technique is NN quantization \cite{gholami2021survey,kim202ibert,zafrir2019q8bert}, where network weights and activations are transformed from 32-bit floating-point representations to integers (typically 8-bit).
Running inference using integer operations has two key advantages.
First, the model size footprint is considerably reduced \textit{e.g.} 8-bit quantization shrinks models by a factor of four.
Second, inference throughput is significantly increased by using more efficient integer-based ``single instruction multiple data'' (SIMD) \cite{patterson2012computer} instructions while improving memory bandwidth utilization, which is typically a bottleneck limiting computational throughput for NNs \cite{quinn-ballesteros-2018-pieces}.

Fundamentally, quantization leads to a quantitative loss of information due to the lowered numerical precision.
As a result, applying integer quantization directly to NN models leads to considerable drop in accuracy \cite{zafrir2019q8bert}.
However, by carefully adjusting the quantization parameters such as the clipping thresholds, the accuracy loss can be significantly reduced, if not eliminated.

The majority of quantization research \cite{gholami2021survey} involve a mix of quantization-aware training (QAT) and post-training calibration techniques with varying complexities to resolve the quantization performance gap.  
Several works \cite{kim202ibert,wang2018pact,aojun2017incrementalnetwork,wang2018pact,raghuraman2018quantizing,louizos2018relaxed,mckinstry2019discovering} detail techniques for QAT as well as approaches where the quantization parameters are optimized using statistics gathered during training.
While these approaches typically close the gap in the quantized model accuracy, they requires access to the training pipeline as well as the training data. 
In addition, these methods are not applicable to black-box models where both training procedures and data are not available.
Also, these methods may be affected by training instabilities, increasing the complexity of the training regimes as described in \cite{raghuraman2018quantizing}.
Post-training approaches such as \cite{tensorrt2017,bhandare2019efficient8bit} require calibration techniques on selected datasets.
For example, in \cite{tensorrt2017} KL-divergence \cite{kl1951} between the unquantized and quantized activations on each layer was used to tune the quantization clipping thresholds. 
Special care needs to be taken when selecting a calibration dataset; as it needs to be diverse enough but yet task specific.
In certain cases this leads to low accuracy, or even unpredictable behaviour, if the run-time input deviates from the calibration dataset.

Two methods that share our high-level goals of eliminating the need for training datasets are introduced in \cite{nagel2019datafree,cai2020zeroq}.
These methods are implemented with CNN-based \cite{gehring2017convolutional} networks, and are used for image classification and object detection tasks. \cite{nagel2019datafree} reduces the quantization error by re-scaling the weights of consecutive CNN layers while taking advantage of the equivariance property of the piece-wise linear ReLU function.
\cite{cai2020zeroq}, on the other hand, tunes the quantization parameters using synthetic data generated utilizing mean and variance statistics obtained from the batch normalization layers of the model itself. 
While both methods are applicable for mainly CNN-based networks, our algorithm is considerably simpler to implement and targets transformers \cite{vaswani2017attention}; particularly SOTA NLP networks with BERT-like \cite{devlin2018bert,liu2019roberta} pre-trained representations.


In this work, we present a method that utilizes the Interquartile Range (IQR) \cite{tukey1977exploratory,rousseeuw1993alternatives}, which is a measure of statistical dispersion, to clip the activations dynamically during inference time.
Our method ensures that at least 75\% of the token-wise extreme activations are not modified, while leaving the remaining 25\% to be statistically modified as outliers, leading to a robust behaviour while considerably improving quantization accuracy.
Our method works for any transformer-based ``trained'' model and does not require any form of training or calibration.
Overall, our contributions can be summarized as follows:
\begin{itemize}  
    \item We propose a novel ``ready-to-use'' inference-time dynamic quantization method that does not require sophisticated re-training/fine-tuning and additional calibration strategies.
    \item Empirically our proposed model demonstrates both effectiveness and robustness on several different NLP benchmark tasks.
    \item Further, contrary to prior work, experiments suggest that our proposed method works both for monolingual and multilingual transformer architectures out-of-the-box.
\end{itemize}



\section{Methodology}
\subsection{Backgound}
Existing approaches to speeding up inference for Transformers mostly focus on General Matrix Multiply (GEMM) operations.
Fast GEMM implementations routinely use GPU and CPU specific SIMD instructions, to execute many multiplications and additions in parallel.
They also optimize memory access patterns to make the best use of available memory bandwidth.  
Integer quantization speeds up the GEMM operations by increasing the amount of data transferred with each memory transaction.
They also take advantage of denser SIMD instructions.
For example, 8-bit quantization packs four times the data per memory transaction compared to 32-bit floating point values.
Many CPUs also support 8-bit SIMD multiplication operations, providing faster as well as cost-effective computation.

\subsubsection{Uniform Quantization}
Dynamic quantization for inference quantizes activations at run time.
The model weights are typically quantized once ahead of execution.
Let $\mathcal{M}\in\mathbb{R}^{m\times n}$ be a matrix of either an activation or parameter weights.
The quantization scale (QS) is obtained as:\vspace{-1em}
\providecommand{\abs}[1]{\lvert#1\rvert}
\begin{equation}
    \label{eqn:qs}
    \mathrm{QS} = \max_{\substack{\forall i \in \{1,\dots,m\} \\ \forall j \in \{1,\dots,n\}}}\abs{\mathcal{M}(i,j)}.
\end{equation}
The matrix $\mathcal{M}$ is then quantized to $\bar{\mathcal{M}}\in \mathbb{Z}^{m\times n}$ as follows:\vspace{-1em}
\begin{equation}
    \label{eqn:q}
   \bar{\mathcal{M}} = \mathrm{int}\left(\frac{2^b/2-1}{\mathrm{QS}} \mathcal{M} \right),
\end{equation}
where $b$ is the number of integerization bits, typically 8, and the function $\mathrm{int}$ is the element-wise integer conversion operator; e.g. a floor function.
The reason for the subtraction by $1$ in (\ref{eqn:q}) is to ensure that the quantization range is equally spread around zero. In the case of 8 bits, the range becomes $\pm 127$.
This formulation also results in a symmetric form of uniform quantization, where the quantization is evenly split around zero.
This can be modified by adding a zero-shift resulting in an asymmetric quantization \cite{raghuraman2018quantizing}, which may particularly be useful for certain activation functions such as ReLU \cite{relu01} and GELU \cite{gelu}.
While non-uniform quantization \cite{gholami2021survey} has been explored to better capture weight and activation distribution with variable step sizes, uniform quantization leads to more efficient implementation on current hardware such as GPUs and CPUs with acceptable accuracy.
Once matrices are quantized, GEMM operations can be performed using integer arithmetic allowing the use of fast SIMD instruction sets.

Quantization lowers numerical precision which leads to loss of information.
Examining (\ref{eqn:qs}) shows how the QS can increase precision errors if it takes extreme values that largely deviate from the majority activations.
Therefore, the activation tensor must be clipped to reduce the quantization error; however, excessive clipping can lead to distortions in the activation which also leads to drops in accuracy.

In the following section, we will outline a method that chooses better QS values for each activation tensor dynamically during inference, without any modification to the training pipeline or any requirement for calibration procedures.

\subsection{Interquartile Range Clipping}
If we consider the extreme values in the activations as outliers in a distribution, there is a substantial amount of research for identifying outliers \cite{ben2005outlier,hodge2004survey}.
Our solution makes use of a low complexity univariate statistical-based method for outlier detection referred to as the Interquartile Range (IQR) method originally proposed by Tukey \cite{tukey1977exploratory}.

IQR is also considered a robust statistical measure \cite{rousseeuw2011robust} of the data spread, with the notion of robustness being defined using the concept of a \textit{breakdown point} \cite{rousseeuw1993alternatives,rousseeuw2011robust}.
The breakdown point is the minimum number of data that can be arbitrarily replaced while keeping the statistical measure bounded.  The sample mean and variance have a 0 breakdown point, meaning that these measures are changed by even a single outlier; on the other hand, the IQR has a 25\% breakdown point, making it a stable measure even if up to 25\% of the data are outliers.

We introduce an algorithm that effectively uses IQR to clip outliers from an activation tensor which consequently improves the selection of the quantization scale as in (\ref{eqn:qs}).
It is worth noting that a direct implementation of the IQR method is too slow as it uses a sorting operation in order to identify the quartiles on the data.
The complexity of a naive implementation would be $\mathcal{O}(N\log{}N)$ where $N$ is the number of elements of the activation tensor.
In the case of BERT-like models, $N=L\times H$, where $L$ is the sequence length and $H$ is the hidden dimension; \textit{e.g.} for BERT-Large, $N=512\times1024$.
To lower this complexity, we obtain the IQR clipping threshold from a reduced set formed by taking the maximums, in absolute sense, along the $H$ dimension.
We will refer to this algorithm as the Token-Maximums IQR (TM-IQR) clipping.
The resulting complexity of the IQR clipping becomes $\mathcal{O}(N+ L \log{}L)$.
Our experiments show that adding this form of IQR clipping slows inference by less than 2\%, which is negligible considering the resulting accuracy gains.

\begin{algorithm}
\small
\caption{Activation clipping using TM-IQR}\label{alg:iqr}
\begin{algorithmic}[1]
\Statex Input: Activation tensor $\mathcal{A} \in \mathbb{R}^{L\times H}$
\Statex $\mathcal{L} \gets \{1,2,\dots,L\}$
\Statex $\mathcal{H} \gets \{1,2,\dots,H\}$
\State \label{alg:iqr:m} $\displaystyle M(i) \gets \max_{\forall j \in \mathcal{H}}\abs{\mathcal{A}(i,j)}$, $\forall i \in \mathcal{L}$
\State \label{alg:iqr:srt} $M \gets \text{sort}(M)$
\State $q1 \gets \text{first-quartile}(M)$
\State $q3 \gets \text{third-quartile}(M)$
\State \label{alg:iqr:thr} $t \gets q3 +1.5(q3-q1)$
-\State \label{alg:iqr:ct} $\mathcal{A}(i,j) \gets \min(\mathcal{A}(i,j), t),~~~~\scriptstyle \forall(i,j) \in \mathcal{L}\times\mathcal{H}$
\State \label{alg:iqr:cb} $\mathcal{A}(i,j) \gets \max(\mathcal{A}(i,j), -t),~\scriptstyle \forall(i,j) \in \mathcal{L}\times\mathcal{H}$
\Statex Return: $\mathcal{A}$
\end{algorithmic}
\end{algorithm}

\noindent Algorithm~\ref{alg:iqr} outlines the basic procedure of our TM-IQR clipping.
In Line~\ref{alg:iqr:m} we compose the set of token-maximum activations in the absolute sense.
Essentially, we are reducing the set of activations to a smaller representative set that contains the top outliers of the larger set.
Lines \ref{alg:iqr:srt} to \ref{alg:iqr:thr} compute the IQR threshold $t$ which is then used to clip the entire activation tensor in lines \ref{alg:iqr:ct} and \ref{alg:iqr:cb}.
The value $1.5$ in line \ref{alg:iqr:thr} is commonly referred to as the IQR scale.
It was historically proposed by Tukey \cite{tukey1977exploratory} as a level to detect outliers.
It is possible to attempt to fine-tune this value, however we chose to use the historical value without tuning in line with the objective of our paper.

It is important to note that the TM-IQR algorithm assigns a dynamic clip value for each activation tensor as opposed to using a fixed value for all run-time inference.
Unlike fixed clipping tuned by training datasets, we expect TM-IQR clipping to be applied in a zero-shot approach across multiple tasks while maintaining reasonable empirical accuracy.
This is due to the fact that our clipping strategy guarantees that at least 75\% of the row-wise extreme activations are not impacted by it, while a fixed clipping method does not offer such guarantees for all types of input, as is the case when the input is not very aligned with training data.
This has the important effect of limiting the distortion error, which occurs when quantizing activations with excessive clipping.

\section{Experimental Setup}
 Our run-time inference engine, implemented in C++, 
supports both FP32 and optimized 8-bit integer quantized inference (I8).
We quantize model weights at load-time and dynamically quantize activations at run-time.
The TM-IQR technique is a straightforward modification with a negligible impact on inference speed, as shown in Table~\ref{tab:iqrspeed}.

\begin{table}
\small
\centering
\begin{tabular}{l c c c | c}
\textbf{CPU} & \textbf{Precision} & \textbf{Method} & \textbf{Batch} & \textbf{WPS} \\
\hline
\hline
Xeon 8260 & int8 & none & 48 & 29005 \\
Xeon 8260 & int8 & IQR & 48 & 28640 \\
V100 & fp16 & none & 128 & 71998 \\
\end{tabular}
\caption{IQR throughput cost in WPS (words per sec) averaged over 4 runs. Each input is 512 tokens.  48 core Xeon 8260 and V100 speed included for reference. }
\label{tab:iqrspeed}
\end{table}

\subsection{TM-IQR} TM-IQR can be applied to the activations before each quantized GEMM operation.
However, we found that the second feed-forward GEMM, henceforth referred to as FF2, contributes the majority of the quantization error.
The input dimension of FF2 is very wide, $4\times{H}$, providing more of a chance for saturation and integer numerical instability to accumulate.
In addition, the input to FF2 constitutes the activations of either ReLU or a GELU non-linearities.
The range of such activation functions is unbounded on the positive side, which further increases the chance of saturation.
Therefore, we found it most effective to apply the TM-IQR to the input activations of the FF2 GEMM operation.

\subsection{Tasks} We test our proposed method on GLUE \cite{DBLP:journals/corr/abs-1804-07461} and 2 popular question answering (QA) tasks: Natural Questions (NQ) \cite{47761} and TyDI \footnote{Note that TyDI is multilingual among 11 typologically diverse languages.} \cite{DBLP:journals/corr/abs-2003-05002}. We train all our tasks using the publicly available \cite{wolf2019huggingface}. For GLUE tasks, we run 5 seeds with hyper-parameters using HuggingFace's defaults for BERT while tuning the learning rate for RoBERTa (refer to \ref{sec:glueApp} for more details). For QA tasks, we follow \cite{alberti2019bert,DBLP:journals/corr/abs-2003-05002}. 
Our underlying pre-trained language model for GLUE is both BERT (cased) \cite{devlin2018bert} and RoBERTa \cite{liu2019roberta}, while for QA, we used XLM-R \cite{conneau2019unsupervised}. 
Note our method \textit{does not need} any fine-tuning once this step is done and models are obtained.


\section{Results}

Since our method does not modify the training pipeline or tune the quantization parameters on training sets, we compare our results directly to the FP32 numbers.
We are not expecting our method to outperform FP32 but rather to reduce the negative effect of quantization while keeping its speed as well as simplifying the model deployment process.


\begin{table}
\small
\centering
\begin{tabular}{l | c c  c}
\textbf{Task} & \textbf{FP32}  & \textbf{I8} & \textbf{TM-IQR} \\
\hline
\hline
XLM-R-base TyDI & 67.7 &  62.9 & \textbf{67.0}\\
XLM-R-large TyDI & 68.8 & 66.8 & \textbf{68.4} \\
\hline
XLM-R-base NQ & 54.6 & 48.0 & \textbf{53.4}\\
XLM-R-large NQ & 56.6 &  53.3 & \textbf{56.1}\\
\end{tabular}
\caption{Question Answering performance.}
\label{tab:qa}
\end{table}

\subsection{Question Answering} On TyDI and NQ (Table~\ref{tab:qa}), TM-IQR clearly recovers most of the performance lost to dynamic quantization and is superior to I8 by 1 point on average.
Similar to GLUE, TM-IQR still performs well with the I8 drop being the highest.



\begin{table}[!h]
\parbox{0.98\columnwidth}{
\small
\centering
\begin{tabular}{ l| l l l }
\textbf{Task} & \textbf{FP32} & \textbf{I8}  & \textbf{TM-IQR} \\
\hline \hline
\multicolumn{1}{l|}{\textbf{BERT-base-cased}} \\ 
MNLI          & 83.7 (0.2)  & 82.3 (0.5) & \textbf{83.5} (0.3) \\
MNLI-MM       & 84.1 (0.1)  & 82.9 (0.2) & \textbf{83.8} (0.2) \\
CoLA          & 58.0 (1.4)  & 48.3 (0.9) & \textbf{57.7} (1.6) \\
SST-2         & 92.3 (0.3)  & \textbf{92.1} (0.2) & 92.0 (0.4) \\
MRPC          & 88.5 (1.2)  & \textbf{88.8} (1.6) & 88.5 (1.5) \\
STS-B         & 88.3 (0.8)  & 87.7 (0.8) & \textbf{88.1} (0.8) \\
QQP           & 87.4 (0.1)  & 86.2 (0.3) & \textbf{87.2} (0.2) \\
QNLI          & 90.8 (0.2)  & 90.3 (0.1) & \textbf{90.5} (0.2) \\
RTE           & 64.6 (1.0)  & 63.9 (1.0) & \textbf{64.9} (1.6) \\
\multicolumn{1}{c|}{\textbf{Average}}   & 82.0        & 80.3       & \textbf{81.8}       \\
\hline\hline
\multicolumn{1}{l|}{\textbf{BERT-large-cased}} \\ 
MNLI          & 86.4 (0.1)  & 86.0 (0.2) & 86.0 (0.1) \\
MNLI-MM       & 86.5 (0.2)  & 86.3 (0.1) & 86.3 (0.2) \\
CoLA          & 62.9 (0.8)  & 60.6 (1.5) & \textbf{62.1} (1.2) \\
SST-2         & 93.3 (0.5)  & 92.8 (0.7) & \textbf{92.9} (0.4) \\
MRPC          & 90.5 (0.5)  & 89.6 (0.9) & \textbf{90.5} (0.7) \\
STS-B         & 89.6 (0.6)  & 87.4 (1.2) & \textbf{89.1} (0.3) \\
QQP           & 88.3 (0.2)  & 88.1 (0.1) & 88.1 (0.1) \\
QNLI          & 92.4 (0.1)  & 91.9 (0.1) & \textbf{92.2} (0.2) \\
RTE           & 69.8 (1.4)  & 64.0 (2.0) & \textbf{68.5} (1.7) \\
\multicolumn{1}{c|}{\textbf{Average}} & 84.4        & 83.0       & \textbf{84.0}       \\
\hline \hline
\multicolumn{1}{l|}{\textbf{RoBERTa-base}} \\ 
MNLI          & 87.0 (0.1)  & 85.8 (0.3) & \textbf{86.1} (0.1) \\
MNLI-MM       & 87.1 (0.1)  & 85.8 (0.2) & \textbf{86.1} (0.1) \\
CoLA          & 53.7 (1.9)  & 22.7 (4.7) & \textbf{50.8} (1.8) \\
SST-2         & 93.9 (0.2)  & 93.4 (0.4) & \textbf{93.5} (0.3) \\
MRPC          & 78.6 (2.7)  & 77.2 (1.1) & \textbf{78.4} (2.0) \\
STS-B         & 87.1 (0.8)  & 69.6 (0.8) & \textbf{85.5} (0.8) \\
QQP           & 88.3 (0.1)  & 87.2 (0.2) & \textbf{87.6} (0.1) \\
QNLI          & 92.5 (0.1)  & 90.1 (1.9) & \textbf{91.4} (0.3) \\
RTE           & 68.0 (2.1)  & 64.7 (2.5) & \textbf{67.4} (2.8) \\
\multicolumn{1}{c|}{\textbf{Average}}   & 82.0        & 75.2       & \textbf{80.8}       \\
\hline\hline
\multicolumn{1}{l|}{\textbf{RoBERTa-large}} \\ 
MNLI          & 90.6 (0.0)  & 90.0 (0.2) & \textbf{90.3} (0.1) \\
MNLI-MM       & 90.0 (0.3)  & 89.6 (0.1) & 89.6 (0.2) \\
CoLA          & 63.5 (0.6)  & 63.1 (1.3) & \textbf{63.4} (0.6) \\
SST-2         & 96.3 (0.4)  & 95.8 (0.2) & 95.8 (0.4) \\
MRPC          & 89.6 (0.4)  & \textbf{90.1} (0.7) & 88.7 (0.8) \\
STS-B         & 91.8 (0.1)  & 91.2 (0.2) & \textbf{91.4} (0.2) \\
QQP           & 89.8 (0.1)  & 89.4 (0.2) & 89.4 (0.2) \\
QNLI          & 94.6 (0.2)  & 94.1 (0.3) & 94.1 (0.3) \\
RTE           & 77.7 (2.0)  & 76.0 (5.1) & \textbf{77.3} (1.5) \\
\multicolumn{1}{c|}{\textbf{Average}} & 87.1        & 86.6       & \textbf{86.7}       \\
\end{tabular}
\caption{The TM-IQR clipping algorithm on GLUE tasks with three computational modes, 32-bit floating-point (FP32), 8-bit quantization (I8) and our algorithm TM-IQR. Metric values are mean and standard deviation (in parenthesis)  over 5 seeds.}
\label{tab:glue}
}
\end{table}

\subsection{GLUE} Table ~\ref{tab:glue} shows that TM-IQR is robust with an overall average score drop, compared to FP32, by \textit{only} 0.2\% for BERT-base, 0.5\% for BERT-large, 1.2\% for RoBERTa-base and 0.4\% for RoBERTa-large.
For all 4 pretrained models, TM-IQR wins on average.  Even when TM-IQR does not outperform I8, the loss is relatively small.
Interestingly, TM-IQR does well for cases where I8 drop is large, \textit{e.g.} CoLA and RTE for all models and STS-B for RoBERTa-base.

\section{Conclusion}
We show that BERT-like models can be quantized to 8-bit integers with good accuracy without the need to modify training procedures or add extra data sets for parameter calibration.
We present a robust statistically-based algorithm that dynamically adjusts the quantization clipping to maintain reasonable accuracy.
Our empirical results demonstrate the effectiveness of our method on a number of NLP monolingual and multilingual tasks, trained on both base and large size BERT-like models.

\bibliography{main}
\bibliographystyle{acl_natbib}

\appendix

\section{Evaluation on GLUE Task}
\label{sec:glueApp}

For GLUE experiments we use the publicly available open-source library \texttt{PyTorch-Transformers} \cite{wolf2019huggingface}.
We report the standard metric on each task, specifically:
Accuracy is used for MNLI, MNLI-MM (mismatch) \cite{williams-etal-2018-broad}, SST-2 \cite{sst}, QNLI \cite{qnli}, and RTE \cite{rte}.
Mathews correlation coefficient is used for CoLA \cite{cola}.
F1 is used for MRPC \cite{mrpc} and QQP \cite{qqp}.
Finally, Pearson correlation coefficient is used for STS-B \cite{stsb}, 
For BERT models, We use the default hyper-parameters provided by the HuggingFace's library, specifically the learning rate is $2.\times 10^{-5}$, the batch-size is $32$  and the fine-tuning epochs is $3$, except for MRPC where the the fine-tuning epochs is $5$.  For RoBERTa models, we tuned the learning rate in $[5e-7, 2e-6]$ for best devset results on FP32 evaluations, in addition we increase the epochs to $6$ for the two large datasets, MNLI and QQP, and to $12$ for the rest of the tasks.
Similarly to \cite{kim202ibert} we exclude WNLI \cite{levesque2012wnli} since it showed unstable results even on FP32 due to its small dataset.

\end{document}